\documentclass[runningheads]{llncs}
\usepackage{graphicx}
\usepackage{color}

\makeatletter
\renewcommand*{\@fnsymbol}[1]{\ensuremath{\ifcase#1\or *\or \dagger\or \ddagger\or
   \mathsection\or \mathparagraph\or \|\or **\or \dagger\dagger
   \or \ddagger\ddagger \else\@ctrerr\fi}}
\makeatother

\begin{document}
\title{DCVQE: A Hierarchical Transformer for Video Quality Assessment}
%
%\titlerunning{Abbreviated paper title}
% If the paper title is too long for the running head, you can set
% an abbreviated paper title here
%
 \author{Zutong Li\thanks{Work done when Z. Li was at Weibo. Z. Li is currently with Microsoft.} \and Lei Yang\thanks{Corresponding author.} }

%Z. Li is now with Microsoft
%

% First names are abbreviated in the running head.
% If there are more than two authors, 'et al.' is used.
%
\institute{Weibo R\&D Limited, USA\\
\email{\{zutongli0805, trilithy\}@gmail.com}}

% \institute{Weibo R\&D Limited, USA}
%
\maketitle              % typeset the header of the contribution
\begin{abstract}
The explosion of user-generated videos stimulates a great demand for no-reference video quality assessment (NR-VQA). Inspired by our observation on the actions of human annotation, we put forward a Divide and Conquer Video Quality Estimator (DCVQE) for NR-VQA. Starting from extracting the frame-level quality embeddings (QE), our proposal splits the whole sequence into a number of clips and applies Transformers to learn the clip-level QE and update the frame-level QE simultaneously; another Transformer is introduced to combine the clip-level QE to generate the video-level QE. We call this hierarchical combination of Transformers as a Divide and Conquer Transformer (DCTr) layer. An accurate video quality feature extraction can be achieved by repeating the process of this DCTr layer several times. Taking the order relationship among the annotated data into account, we also propose a novel correlation loss term for model training. Experiments on various datasets confirm the effectiveness and robustness of our DCVQE model.

\end{abstract}

\section{Introduction}

Recent years have witnessed a significant increase in user-generated content (UGC) on social media platforms like Youtube, Tiktok, and Weibo. Watching the UGC videos on computers or smartphones has even become part of our daily life. This trend stimulates a great demand for automatic video quality assessment (VQA), especially in popular video sharing/recommendation services. 

UGC-VQA, also known as blind or No-Reference video quality assessment (NR-VQA), aims to evaluate in-the-wild videos without the corresponding pristine reference videos. Usually, UGC videos may suffer from complex distortions due to the diversity of capturing devices, uncertain shooting skills, compression, and poor editing process. Although many excellent algorithms have been proposed to evaluate video quality, it remains a challenging task to assess the quality of UGC videos accurately and consistently.

\begin{figure}[t]
\begin{center}
   \includegraphics[width=.58\columnwidth]{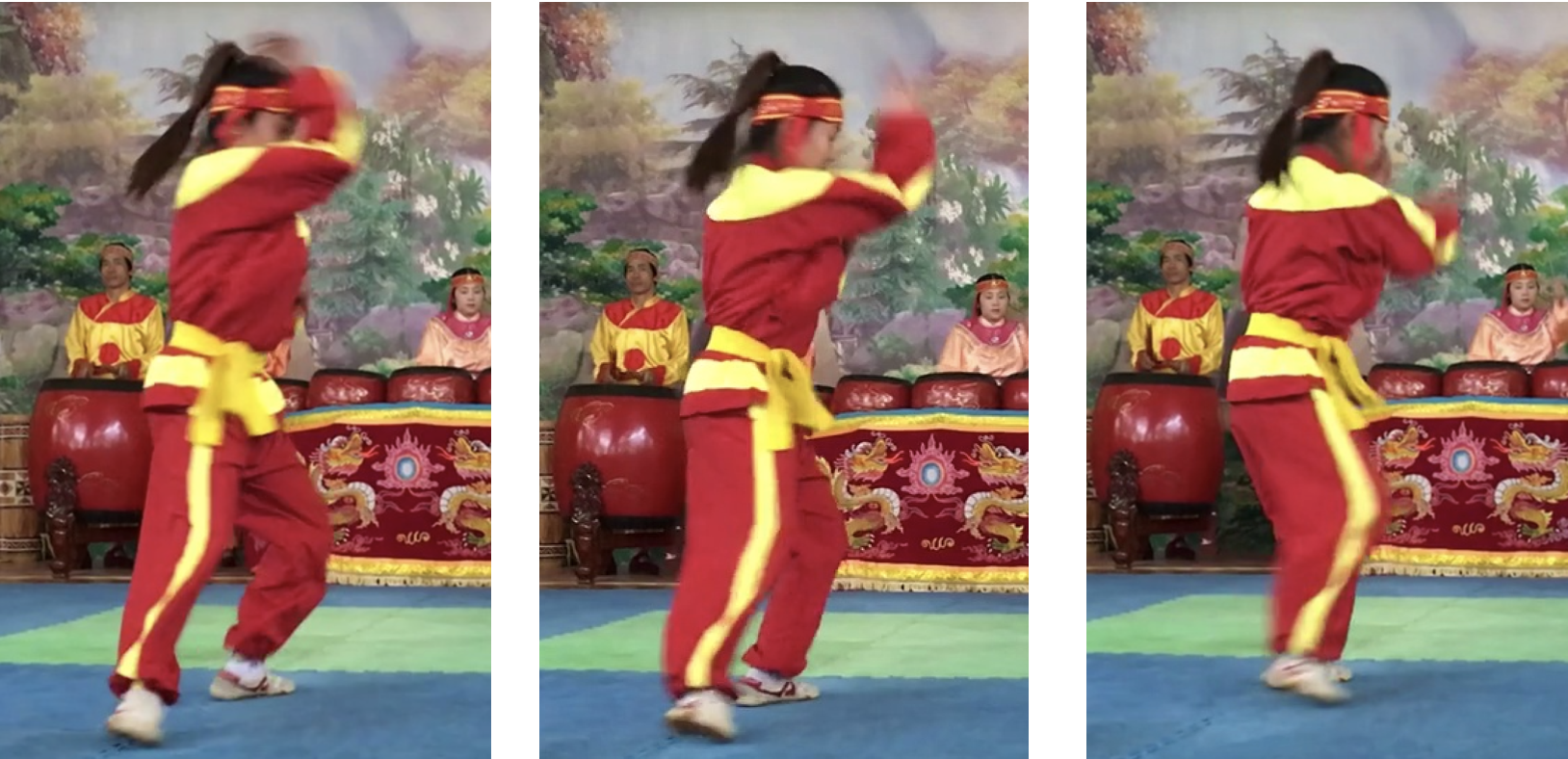}
\end{center}
   \caption{Three consecutive frames from Video B304 in LIVE-VQA dataset. An overall high mean opinion score (MOS) of 91.73 was annotated.}
   %Almost all frames contain motion blur (see hand area). While watching these frames continuously in the video, the movement of this actress is actually smooth, and as a result, an overall high mean opinion score (MOS) of 91.73 was annotated.}
\label{fig:0}
\end{figure}

Besides the frame-level image information, temporal information is regarded as a critically important factor for video analysis tasks. Although many image quality assessment (IQA) models \cite{NIQE,ILNIQE,AMittal_BRISQUE,WXue_GMLOG,HIGRADE,FRIQUEE,CORNIA,HOSA,PAQ2PIQ} can be applied to VQA base on a simple temporal pooling process \cite{ZTuTemporalPooling}, these models may not work very robustly because of the absence of proper time sequence aggregation. For instance, Fig. \ref{fig:0} shows three consecutive frames extracted from Video B304 in LIVE-VQA \cite{LIVEVQC} dataset. As seen, the motion blur distortion appears on the actress’s hand area. These frame images are most likely to be recognized as middle, even low quality when applying a sophisticated IQA method to them individually. However, the quality of this video was labeled as high by human annotators, because a very smooth movement of the actress can be observed when playing the video stream. RNNs \cite{VIDEO_RNN1,VIDEO_RNN2,VSFA} and 3D-CNNs \cite{3DCNN,I3D,SLOWFAST,X3D,SMALLBIG} are potential models to integrate spatial and temporal information for NR-VQA. Though these algorithms perform well on many datasets, the difficulty of parallelizing RNN models and the non-negligible computational cost of 3D-CNNs make them infeasible to be applied to many internet applications that require quick responses, just like the online video sharing/recommendation services.

On the other hand, research on NR-VQA is often limited by the lack of sufficient training data, due to the tedious work to label the mean opinion score (MOS) for each video. Among the publicly available datasets, MCL-JCV \cite{MCL_JCV}, VideoSet \cite{VIDEOSET}, UGC-VIDEO \cite{UGC_VIDEO}, CVD-2014 \cite{Nuutinen_CVD}, LIVE-Qualcomm \cite{Ghadiyaram_LIVEQ} are generated in lab environments, while KoNViD-1k \cite{KONVID1K}, LIVE-VQC \cite{LIVEVQC}, YouTube-UGC \cite{YOUTUBEUGC} and LSVQ \cite{PATCHVQ} are collected in-the-wild. As mentioned above, one could consider VQA as a temporal extension of IQA task, thus can apply some frame-level distortion and ranking processes \cite{RANKIQA}  to augment the small video datasets for algorithm development. However, in-the-wild videos are usually hard to synthesize, since they may suffer from compound distortions which cannot be exactly parameterized as a combination of certain distortion cases. Recently, a large-scale LSVQ dataset \cite{PATCHVQ}, which contains 39,075 annotated videos with authentic distortions is released for public research. We know that not many studies have been conducted so far based on this new dataset.

Additionally,  most previous works take L1 or L2 loss \cite{PAQ2PIQ,PATCHVQ,METAIQA,VSFA,RAPIQUE} as the optimization criterion for model training. Since these criteria to some extent ignore the order relationship of quality scores of the training samples, the trained model may be not stable to quantify the perceptual differences between the videos with similar quality scores. For example, in our research we find that many existing NR-VQA models work well to identify both high and low quality videos, but struggle to distinguish the videos with middle quality scores. How to effectively quantify the difference between samples with similar perceptual scores therefore becomes the key to the success of a NR-VQA model. 

To address the above problems, in this paper we put forward a new Divide and Conquer Video Quality Estimator (DCVQE) model for NR-VQA. We summarize our contributions as follows: (1) Inspired by our observation on the actions of human annotation, we propose a Divide and Conquer Transformer (DCTr) architecture to extract video quality features for NR-VQA. Our algorithm starts from extracting the frame-level quality representations. Regarded as a divide process, we split the input sequence into a number of clips and apply Transformers to learn the clip-level quality embeddings (QE) and update the frame-level QE simultaneously. Subsequently, a conquer process is conducted by using another Transformer to combine the clip-level QE to generate a video-level QE. After stacking several DCTr layers and topping with a linear regressor, our DCVQE model can be constructed to predict the quality value of the input video. 
(2) By taking the order relationship of the training samples into account, we propose a novel correlation loss to bring an additional order constraint of video quality to guide the training. Experiments indicate that the introduction of this correlation loss can consistently help to improve the performance of our DCVQE model. (3) We conduct plenty of experiments on different datasets and confirm that our DCVQE outperforms most other algorithms.%\zt{ and achieves the state-of-the-art for NR-VQA.} 

\section{Related works}

\textbf{Traditional NR-VQA solutions:} Many prior NR-VQA works are ``distortion specific"  because they are designed to identify different distortion types like blur \cite{BLUR}, blockiness \cite{BLOCK}, or noise \cite{NOISE} in compressed videos or image frames. More recent and popular used models are deployed on natural video statistics (NVS) features, which are created by extending the highly regular parametric bandpass models of natural scene statistics (NSS) from IQA to VQA tasks. Among them, successful applications have explored in both frequency domain (BIQI \cite{BIQI}, DIIVINE \cite{DIIVINE}, BLINDS \cite{BLINDS}, BLINDS-II \cite{BLINDS2}) and spatial domain (NIQE \cite{NIQE}, BRISQUE \cite{AMittal_BRISQUE}). V-BLIINDS \cite{VIDEO_BLINDS} combined spatio-temporal NSS with motion coherency models to estimate perceptual video quality. Inter-subband correlations, modeled by spatial domain statistical features in frame differences, were used to quantify the degree of distortion in VIIDEO \cite{VIIDEO}. 3D-DCT was applied on local space-time regions, to establish quality aware features in \cite{3DDCT}. Based on hand-crafted features selection and combination, recent algorithms VIDEVAL \cite{VIDEVAL} and TLVQM \cite{TLVQM} demonstrated outstanding performance on many UGC datasets. The designation of these hand-crafted features is deliberate, though they are hard to be deployed in an end-to-end fashion for NR-VQA tasks.

\textbf{Deep learning based NR-VQA solutions:} Deep neural networks have shown their superior abilities in many computer vision tasks. With the availability of perceptual image quality datasets \cite{CLIVE,KONIQ_10K,TID2013,PAQ2PIQ}, many successful applications have been reported in the past decade \cite{RANKIQA,PAQ2PIQ,BIQA,NIMA,METAIQA}. Combining with a convolutional neural aggregation network, DeepVQA \cite{DEEPVQA} utilized the advantages of CNN to learn spatio-temporal visual sensitivity maps for VQA. Based on a weakly supervised learning and resampling strategy, Zhang et al. \cite{WEAKLYVQA} proposed a general purpose NR-VQA framework which inherited the knowledge learned from full-reference VQA and can effectively alleviate the curse of inadequate training data. VSFA \cite{VSFA} used a pretrained CNN to abstract frame features, and introduced the gated recurrent unit (GRU) to learn the temporal dependencies. Following 
VSFA, MDTVSFA \cite{MDTVSFA} proposed a mixed datasets training method to further improve 
VQA performance. Although the above methods performed well on synthetic distortion datasets, they may be unstable to analyze UGC videos with complex and diverse distortions. PVQ \cite{PATCHVQ} reported a leading performance on the large-scale dataset LSVQ. For a careful  study on the local and global spatio-temporal quality, spatial patch, temporal patch and spatio-temporal patch were introduced in \cite{PATCHVQ}. RAPIQUE \cite{RAPIQUE}, by leveraging a set of NSS features concatenated with learned CNN features, shown the top performance on several public datasets, including KoNViD-1k, YouTube-UGC, and their combination.

\textbf{Transformer techniques in computer vision:} Self-attention mechanism-based Transformer architecture shows its exceptional performance in natural language processing \cite{ATTENTION,BERT}. Recently, many researchers introduced Transformer to solve computer vision problems. ViT \cite{Vit} directly run attention among image patches with positional embeddings for image classification. Detection Transformer (DETR) \cite{DETR} reached a comparable performance with Faster-RCNN \cite{FASTERRCNN} by designing a new object detection systems based on Transformers and bipartite matching loss for direct set prediction. Through conducting contrastive learning on 400 million image-text pairs, CLIP \cite{CLIP}
shown impressive performance to solve different zero-shot transfer learning problems. For IQA tasks, Transformer also shows its powerful strength. Inspired by ViT, TRIQ \cite{TRIQ} connected Transformer with MLP head to predict perceptual image quality, where sufficient lengths of positional embeddings were set to analyze the images with different resolutions. IQT \cite{IQT} achieved outstanding performance by applying Transformer encoder and decoder on the features of reference images and distorted images. Through introducing 1D CNN and Transformer to integrate short-term and long-term temporal information, a recent work LSCT \cite{MM21} demonstrated excellent performance on VQA. Our proposal is also derived from Transformer, inspired by our observation on the actions of human annotation. Experiments on various datasets confirm the effectiveness and robustness of our method.

\section{Divide and Conquer Video Quality Estimator (DCVQE)}

%\lorem{1}
%\subsection{Divide and Conquer Video Quality Estimator (DCVQE)}
Human judgements of video quality are usually content-dependent and affected by their temporal memory \cite{HUMANEFF_1,HUMANEFF_2,HUMANEFF_3,HUMANEFF_4,HUMANEFF_5,HUMANEFF_6,VIDEOSET}. In our investigation, we notice that many human annotators like to give their opinions on the quality of a video after the following two actions: first, watch the video quickly (usually in the fast forward mode) to get an overall impression of its quality, then they may scroll mouse forward and backward to review some specific parts of the video for their final decisions. Inspired by this observation, we propose a hierarchical architecture, dubbed Divide and Conquer Video Quality Estimator (DCVQE) for NR-VQA. Our model is worked by extracting three levels of video quality representations from frames, video clips to whole video sequence progressively and repeatedly, somewhat similar to the reverse processes of human annotation. An additional correlation loss term is also presented to bring an additional order constraint of video quality to guide the training. We find that our method can effectively improve the performance of NR-VQA. We will describe our work in detail in the following paragraphs.

\begin{figure*}
\begin{center}
\includegraphics[width=1\linewidth]{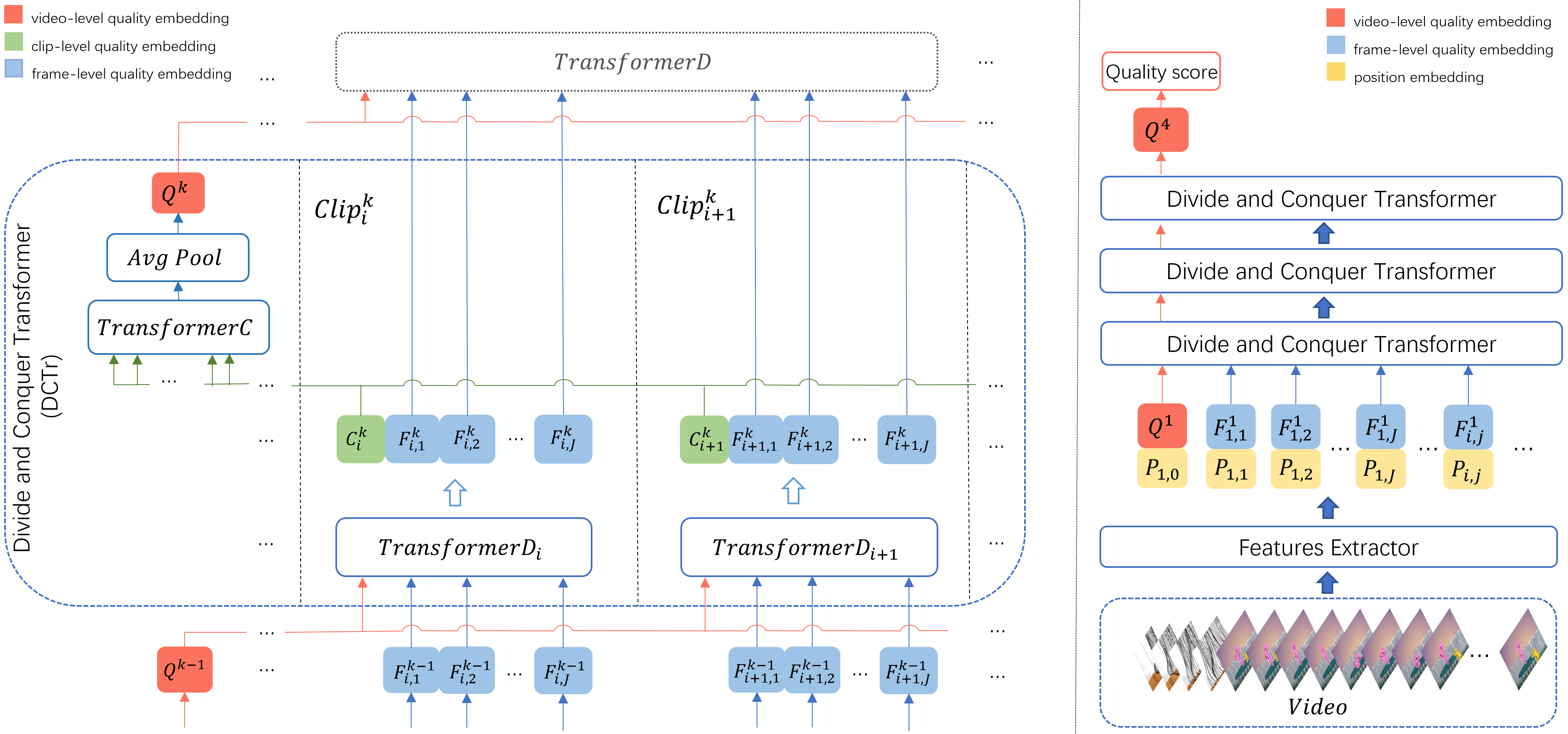}
\end{center}
   \caption{The architectures of the proposed Divide and Conquer Transformer (DCTr) layer (left) and Divide and Conquer Video Quality Estimator (DCVQE) (right).}
\label{fig:1}
\end{figure*}

\subsection{Overall Architecture}
The left side of Fig. \ref{fig:1} represents the architecture of a key video quality analysis layer, Divide and Conquer Transformer (DCTr) in our proposal. In order to simulate the second action of human annotation mentioned above, we split the input sequence into a number of clips, and introduce a Transformer module $TransformerD$ to learn quality representations for each clip. As shown in Fig. \ref{fig:1}, for the $k^{th}$ DCTr layer, we split the whole sequence into $I$ clips, and each clip covers $J$ frame-level quality representations generated by the previous layer. For the $i^{th}$ ($1\!\leq\! i \!\leq\! I$) clip $C^{k}_{i}$, a module $TransformerD_i$ is applied to combine the two levels of quality embeddings (QE) generated by the previous layer, that is, all the $J$ frame-level QE $F^{k-1}_{i,j} (1\!\leq\! j \!\leq\! J)$ and the video-level QE $Q^{k-1}$, to simultaneously learn the clip-level QE $C^k_i$ and update the frame-level QE $F^{k}_{i,j}$ for the current layer. We further simulate the first action of human annotation by integrating the learned clip-level QE to generate a video-level QE. The other Transformer module $TransformerC$ with a topped average pooling layer are proposed to merge all clip-level quality representations $C^k_i$ to predict the video-level QE $Q^k$ for the current layer. In general, our proposed video quality analysis module is constructed in a divide and conquer format, so we coin it as a Divide and Conquer Transformer (DCTr) layer. 

The overall architecture of our DCVQE model is shown in the right side of Fig. \ref{fig:1}. As seen, we stack several DCTr layers (3 layers for our proposal, refer to supplementary material for details about this setting) to extract the final video-level QE $Q^k$ for the input video. The final quality score is predicted by topping this embedding $Q^k$ with a regressor. In practice, to improve the temporal sensitivity of our model, we progressively expand the coverage of video clips in the DCTr as the layers deepen. For example, supposing that each clip in the $k^{th}$ DCTr layer covers $J$ frame-level QE from the previous layer, the coverage of each clip in the $(k\!+\!1)^{th}$ DCTr layer will be  $2J$, which means two neighbor clips of the $k^{th}$ DCTr layer are combined to extract the clip-level QE in the $(k\!+\!1)^{th}$ DCTr layer. For the first DCTr layer, the input frame-level QE $F^{1}_{*, *}$ is generated by our feature extractor described in 
subsection \ref{FeatureEx}, and the input video-level QE $Q^1$ is initialized randomly. Both of them are integrated with positional embeddings $P_{*, *}$ \cite{BERT} for the following processes.

\begin{figure}[t]
\begin{center}
   \includegraphics[width=.89\columnwidth]{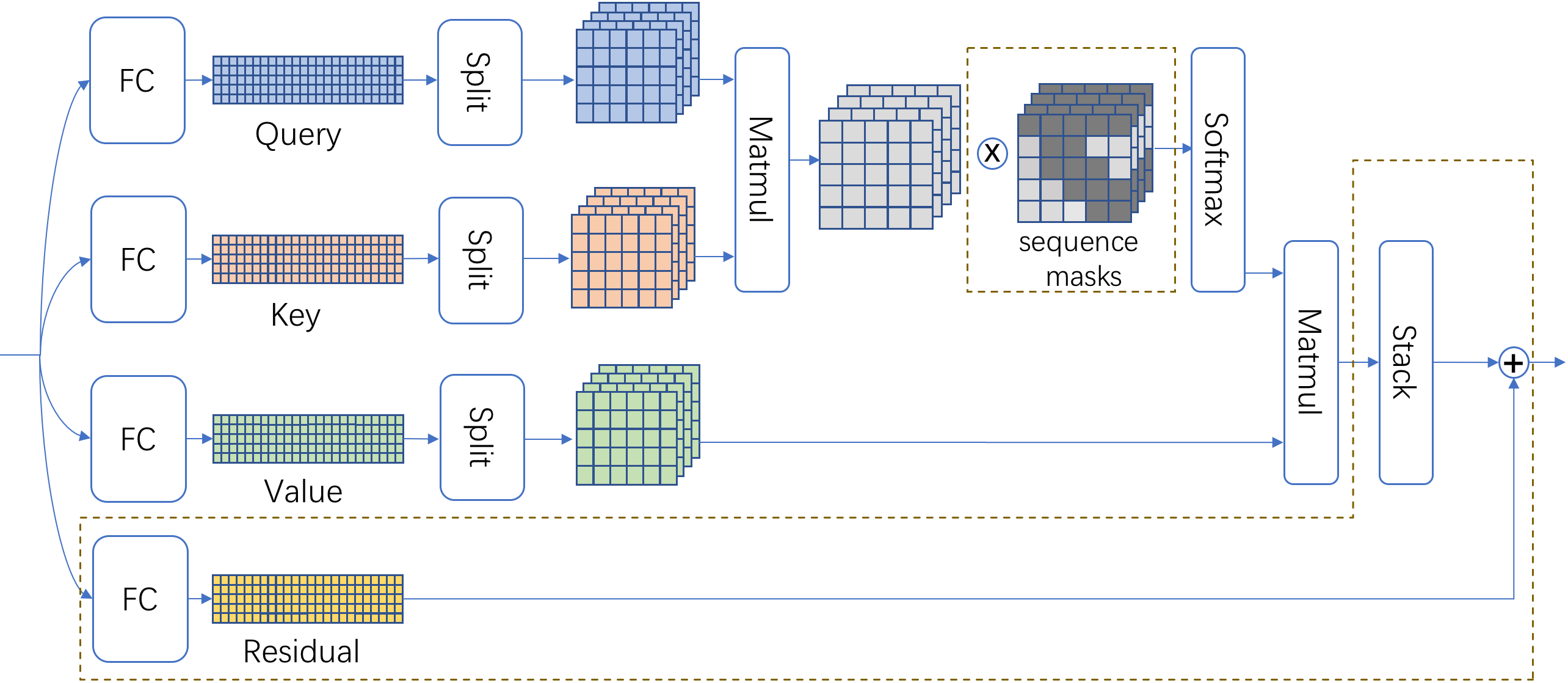}
\end{center}
   \caption{The architecture of $TransformerD$ (include dashed box) and $TransformerC$ (exclude dashed box).}
\label{fig:DNC}
\end{figure}

$TransformerD$ and $TransformerC$ are the two most important modules in our DCVQE. As shown in Fig. \ref{fig:DNC}, they have similar architecture to the classic Transformer \cite{ATTENTION}. Three fully connected layers are used to convert the input features into Query, Key, and Value tensors. Split operations are then adopted to group these three tensors into $H$ heads, respectively. For the input frame-level feature with the shape of $B \!\times\! S \!\times\! D$, where $B$, $S$ and $D$ denote the batch size, sequence length and dimension of the frame embedding respectively, the shapes of its corresponding grouped Query, Key and Value tensors will be  $H \!\times\! B \!\times\! S/H \!\times\! D$. The attention weights can be computed by the dot-product operations between the grouped Query and grouped Key tensors. To make $TransformerD$ module more concerned with the clip-level quality information, except the first position is reserved for the input video-level QE, we mask out the attention weights outside a predefined temporal range by using sequence masks (the dashed square plotted in Fig. \ref{fig:DNC}) and adopt a Softmax layer to normalize the rests. The normalized weights are then applied to the grouped Value tensor to update the quality features. A stack operation is conducted on the head axis to recover the shape of the tensor. Here we also introduce the residual technique \cite{RESNET} to improve the robustness of $TransformerD$ module, so the final output of this module is the addition of the recovered features and residual. $TransformerC$ module is designed to update the video-level QE for each DCTr layer. The difference between $TransformerC$ and $TransformerD$ modules 
is that $TransformerC$ module does not contain sequence masks and residual block, as shown in Fig. \ref{fig:DNC}.

\subsection{Feature Extraction}
\label{FeatureEx}

\begin{figure}[t]
\begin{center}
   \includegraphics[width=0.89\columnwidth]{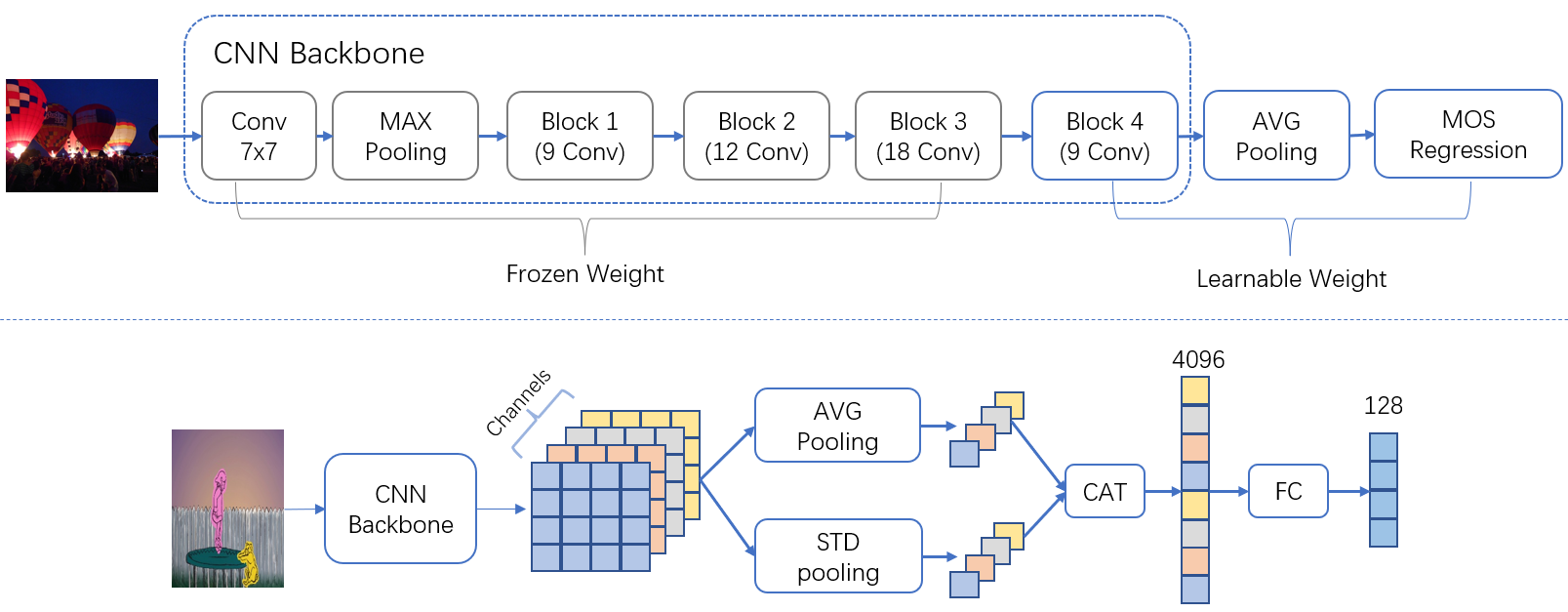}
\end{center}
   \caption{Details of CNN backbone fine-tuning (top) and feature extraction (bottom).}
\label{fig:CNN1}
\end{figure}

% \begin{figure}[t]
% \begin{center}
%   \includegraphics[width=.99\columnwidth]{fig/CNN2.png}
% \end{center}
%   \caption{The feature extraction process}
% \label{fig:CNN2}
% \end{figure}
To avoid the additional distortions that may be caused by applying some preprocessing steps to the input data, such as image resizing and filtering, we take the original full-size image frames as the input of our model. The pretrained Resnet-50 \cite{RESNET} is adopted as the CNN backbone of our feature extractor. To increase the sensitivity of the feature extractor to capture frame-level distortions, we first conduct a typical IQA task to fine-tune the CNN backbone. As shown in the top figure of Fig. \ref{fig:CNN1}, all CNN layers except the last group of CNN blocks are frozen for 
fine-tuning. We find this process can help to transfer the learned knowledge from ImageNet \cite{ImageNet} to our application. It will be verified by our experiments. %\zt{In practice, through topping with an additional average pooling layer and a fully connected layer, the backbone network can be tuned using the simple L1 loss which is calculated between the image quality predictions and their MOS ground truths.}

Based on the fine-tuned backbone, as the bottom figure of Fig. \ref{fig:CNN1} shows, the quality feature of each input frame is then extracted by adopting a concatenation operator to the outputs of global average (AVG) pooling and global standard deviation (STD) pooling processes \cite{VSFA,MDTVSFA}. Since both the dimensions of the outputs of AVG pooling and STD pooling are 2048, the frame-level QE for each frame will be a vector with a dimension of 4096. In practice, refer to \cite{VSFA}, we further add a fully connected (FC) layer to reduce the dimension of the frame-level QE to 128. It is an efficient way to balance the accuracy of frame-level feature representation and overall computational cost. 

\subsection{Correlation loss}
Prior works usually use the following L1 loss to optimize the model:
\begin{equation}
\label{eq1}
L_1=\frac{1}{N} \sum_{n=1}^{N}\left|p_{n}-g_{n}\right|
\end{equation}
where $p_n, n \in [1, N]$ represents the $n^{th}$ predicted MOS, $g_n$ denotes its corresponding ground truth, $N$ is the total number of videos. Mathematically, this criterion to some extent ignores the relative order relationship of quality scores of the training samples in a batch, it sometimes may lead to limited ability to quantify the perceptual differences between the videos with similar quality scores. For example, considering the training processes of  two NR-VQA models based on L1 loss criterion, for two video samples A and B with the MOS ground truths of 7.0 and 7.1, if their quality scores are predicted as $P_{A_1}=7.1$, $P_{B_1}=7.0$ and $P_{A_2}=6.9$, $P_{B_2}=7.0$ by the two models respectively (where $P_{A_1}$ denotes the quality score of video A predicted by model 1, and so forth), the same strengths coming from the L1 losses (0.1) of these two training samples will be contributed to optimize the models correspondingly. Though it is hard to say which model is better than the other, in our opinion, it may be easier to optimize model 2 because of a positive correlation between the ground truths and its loss. Based on this idea, a new correlation loss, aiming to bring an additional order constraint of video quality to guide the training procedure, is proposed as follows:
\begin{small}
\begin{equation}
\label{eq2}
L_{c}=\frac{1}{N} \sum_{n=1}^{N} \max \left(0, -\left(\sum_{m=1}^{N}\left(p_{n}-p_{m}\right)\right)\left(\sum_{m=1}^{N}\left(g_{n}-g_{m}\right)\right)\right)
\end{equation}
\end{small}where $p_m$ represents the prediction of a training sample in a batch with size $N$, and $p_n$ denotes the $n^{th}$ prediction which is picked to compare the ranking order with each element in the prediction set. $g_m$ and $g_n$ stand for the corresponding ground truth scores of $p_m$ and $p_n$, respectively. This loss equation can be considered as that, we take the $n^{th}$ video in the batch as an anchor and compute the positive or negative correlations between predictions and ground truths for this anchor and any other one in the batch. Eq. \ref{eq2} can be further simplified as:
\begin{small}
\begin{equation}
\label{eq3}
L_{c}=\frac{1}{N} \sum_{n=1}^{N} \max \left(0,-N^{2}\left(p_{n}-\sum_{m=1}^{N} \frac{p_{m}}{N}\right) \left(g_{n}-\sum_{m=1}^{N} \frac{g_{m}}{N}\right)\right)
\end{equation}
\end{small}
\begin{equation}
\label{eq4}
L_{c}=N \sum_{n=1}^{N} \max \left(0,-\left(p_{n}-\bar{p}\right)\left(g_{n}-\bar{g}\right)\right)
\end{equation}
where $\bar{p}$ and $\bar{g}$ represent the mean values of predictions and ground truths, respectively. We find that this simplified $L_c$ term is actually equivalent to the Spearman correlation coefficient without normalization. A total loss $L$, which combines the above two loss equations (\ref{eq1}) and (\ref{eq4}), is finally constructed to optimize our DCVQE model as
\begin{equation}
\label{eq6}
L = \alpha \times L_1 + \beta \times L_c
\end{equation}
where $\alpha$ and $\beta$ represent the weights of $L_1$ loss and $L_c$ loss. We will discuss the optimal settings for these two weights in subsection 4.4.

\section{Experiments}
% See \Table{sota} and \Table{ablations}.
% \lorem{5}

\subsection{Datasets} We conduct experiments on four in-the-wild VQA datasets: KoNViD-1k \cite{KONVID1K}, LIVE-VQC \cite{LIVEVQC}, YouTube-UGC \cite{YOUTUBEUGC}, and LSVQ \cite{PATCHVQ}. In these datasets, KoNViD-1k contains 1,200 unique contents with a duration of 8 seconds; LIVE-VQE, the smallest among them, consists of 585 video contents; YouTube-UGC contains 1,500 UGC video clips sampled from millions of YouTube videos; A recently released dataset LSVQ, containing 39,075 video samples with diverse durations and resolutions, is the largest publicly available UGC datasets for research right now. We also follow the suggestion given in \cite{VIDEVAL,INLSA} to select YouTube-UGC dataset as the anchor and map MOS values of KoNViD-1k and LIVE-VQC data onto a common scale to generate an All-Combined dataset for performance evaluation. The typical random 60-20-20 strategy \cite{VSFA} is applied to split data into three sets, i.e. 60\% for training, 20\% for evaluation, and the remaining 20\% for testing.
%\zt{We also implement 60\% training, 20\% evaluation and 20\% testing settings to promote}. 
Especially, for LSVQ we follow the setting in \cite{PATCHVQ} to first generate a Test-1080p set and then conduct the random 80-20 splitting on the rests to generate training and testing sets for experiments.

\subsection{Evaluation metrics} Here we adopt four commonly used criteria, Spearman Rank-Order Correlation Coefficient (SRCC), Kendall Rank-Order Correlation Coefficient (KRCC), Pearson Linear Correlation Coefficient (PLCC) and Root Mean Square Error (RMSE) to measure prediction monotonicity and prediction accuracy. For fair comparisons, we conduct each evaluation 100 times individually and report the median values of these four metrics as their final results.

\subsection{Implementation Details} 
We first fine-tune ImageNet-pretrained Resnet-50 backbone on KonIQ-10k dataset \cite{KONIQ_10K} with random 80-20 splitting and maximum 20 epochs. The CNN backbone with the best performance of SRCC on the test set is selected to construct  the feature extractor for our VQA task. The frame-level features are extracted using this feature extractor, where all frames of each video are considered without sampling. The temporal range of the sequence mask in the proposed $TransformerD$ (see Fig. \ref{fig:DNC}) is empirically set to 15. For a tradeoff between GPU memory, running speed, and performance, we set the maximum length of each input video to 600 (i.e. only the first 600 frames of a video are considered if the video is longer than this maximum length) and the maximum training epoch to 75. An evaluation process will be conducted after each epoch, and the model with the lowest loss on the validation set will be stored.

\subsection{Ablation Studies}
In this subsection, we perform ablation studies to better understand the different components of the proposed DCVQE model.

\textbf{CNN Backbone:} Training of the CNN feature extraction backbone can also be regarded as a typical IQA task. Two series of architectures based on ResNet and ViT backbones are fine-tuned fully or partially in our study. Here we note the partial fine-tuning means that only the last blocks of these models, topped with average pooling and fully connected layers (see Fig. \ref{fig:CNN1}) 
are trainable. We determine that fully fine-tuned backbones act negatively with regard to the complexities of the networks. i.e. deeper structure results in worse performance, and partial fine-tuned backbones usually work better than fully fine-tuned ones for both two series of architectures. We also incorporate a new attention-based IQA architecture PHIQNet \cite{MM21} to our study, and similar performance as partial fine-tuning on ResNet is observed (refer to supplementary material for details). In our application, for a tradeoff between model complexity and performance, also a fair comparison with previous work \cite{VSFA}, we select the partially fine-tuned ResNet-50 as the feature extraction backbone to construct our DCVQE model.

%Test results are listed in Table~\ref{table:IQA_Table}. From rows 1, 3, and 5 of this table, we can see that the fully fine-tuned backbones perform negatively with regard to the complexities of the networks, i.e. a deeper network results in a worse performance. However this phenomenon has been completely changed 
%with the partial fine-tuning processes, as can be seen from rows 2, 4, and 6. It confirms that the partial fine-tuning based transfer learning can effectively improve the sensitivity of the backbone to capture frame-level distortions, compared to the fact that the deeper network trends to overfit the limited training data by the fully fine-tuning procedure. In our application, for a tradeoff between model complexity and performance, we choose the partially fine-tuned ResNet-50 as the CNN backbone of our feature extractor to construct the DCVQE model.

\begin{figure}[t]
\begin{center}
   \includegraphics[width=.89\columnwidth]{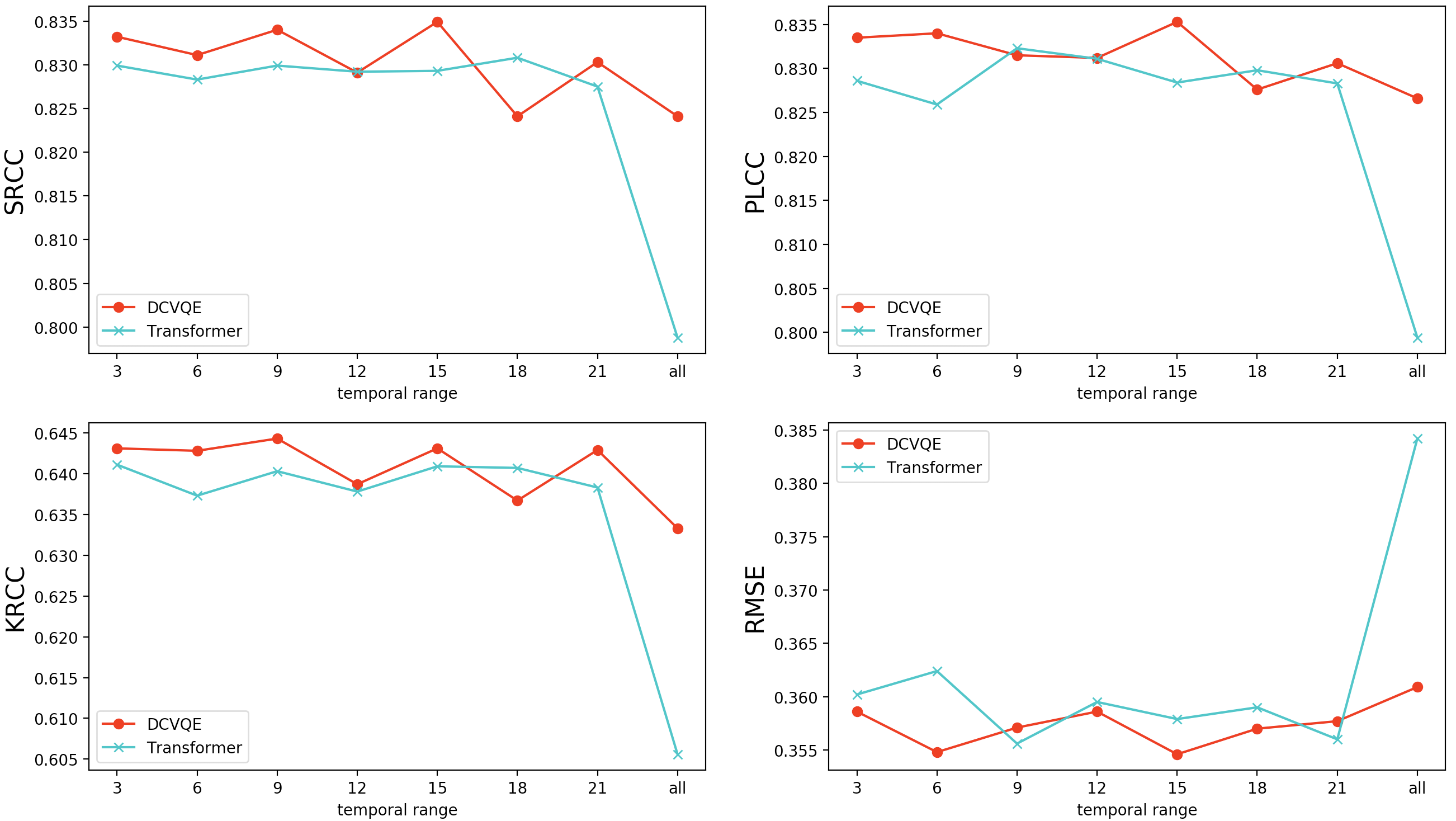}
\end{center}
   \caption{Performance comparisons between a baseline Transformer model and our DCVQE with different settings of the temporal range.}
\label{fig:TemporalRange}
\end{figure}
%\begin{figure}[t]
%\begin{center}
%   \includegraphics[width=.89\columnwidth]{fig/CLLOSS.png}
%\end{center}
%   \caption{Performance comparisons among ‘with correlation loss’ (w CL), ‘without correlation loss’ (w/o CL) and ‘with pairwise ranking loss’ (w PW-RL) under different models and settings on KoNViD-1K dataset.}
%\label{fig:CLLOSS}
%\end{figure}

%\begin{table} \footnotesize
%\begin{center}
%\caption{Performance comparisons of three ResNet architectures with the full and partial fine-tuning strategies.}
%\label{table:IQA_Table}
%\begin{tabular}{lcccc}
%\hline
% & \multicolumn{3}{c}{TVSum}\\
%  Models & SRCC & PRCC &  KRCC & RMSE\\
%\hline
% ResNet-18$_{full}$ &     0.8893 & 0.8798 & 0.6955 & 0.2466   \\
% ResNet-18$_{partial}$ &    0.8888 & 0.8813 & 0.6979 & 0.2494   \\
% ResNet-50$_{full}$ &     0.8507 & 0.8407 & 0.6457 & 0.2889   \\
% ResNet-50$_{partial}$ &    0.9058 & 0.8933 & 0.7168 & 0.2308   \\
% ResNet-101$_{full}$ &    0.8511 & 0.8317 & 0.6365 & 0.2847   \\
% ResNet-101$_{partial}$ &   0.9075 & 0.8962 & 0.7166 & 0.2278   \\ 
%\hline
%\end{tabular}
%\end{center}
%\end{table}

\textbf{Temporal Range of Sequence Mask:} We introduce a group of experiments on KoNViD-1k dataset to study how the temporal range affects the DCVQE model. By directly replacing the proposed DCTr layers with the traditional Transformer layers \cite{ATTENTION}, a baseline Transformer model is constructed for performance comparisons. Test results are plotted in Fig. \ref{fig:TemporalRange}. From these figures, we can see that the best temporal range for the baseline Transformer model is about 9, and the performance of this baseline model slightly decreases as the temporal range increases. Especially if we set the range to “all”, which means that all the frame-level representations are involved in the self-attention processes \cite{ATTENTION} of the Transformer, the performance of this baseline model drops dramatically. Switching to our DCVQE,  we can see that the overall performance stays at a high level. The best performance can be observed when the temporal range reaches 15. The results of the “all” setting for DCVQE, for which one frame only conducts self-attention with all frames in its own clip, also demonstrate the robust performance of our DCVQE model. 

\begin{table} \footnotesize
\begin{center}
\caption{Performance comparisons of different weight combinations of $\alpha$ (for L1 loss term) and $\beta$ (for proposed CL term)  on KoNViD-1k dataset.}
\label{clloss_alpha}
\begin{tabular}{cccccc}
\hline
% & \multicolumn{3}{c}{TVSum}\\
  $\alpha$ & $\beta$ & SRCC & PLCC &  KRCC & RMSE\\
\hline
0.0 & 1.0 & 0.6980 & 0.6956 & 0.5149 & 3.0966\\
%0.1 & 0.9 & 0.7244 & 0.6541 & 0.5329 & 0.6474\\
%0.2 & 0.8 & 0.8156 & 0.8148 & 0.6244 & 0.6421\\
0.3 & 0.7 & 0.8259 & 0.8264 & 0.6360 & 0.3924\\
%0.4 & 0.6 & 0.8292 & 0.8298 & 0.6418 & 0.3594\\
0.5 & 0.5 & 0.8349 & 0.8353 & 0.6431 & 0.3546\\
%0.6 & 0.4 & 0.8365 & 0.8352 & 0.6485 & 0.3533\\
\textbf{0.7} & \textbf{0.3} & \textbf{0.8382} & \textbf{0.8375} & \textbf{0.6500} & \textbf{0.3515}\\
%0.8 & 0.2 & 0.8360 & 0.8399 & 0.6497 & 0.3485\\
%0.9 & 0.1 & 0.8347 & 0.8335 & 0.6420 & 0.3507\\
1.0 & 0.0 & 0.8278 & 0.8306 & 0.6396 & 0.3623\\
\hline
\end{tabular}
\end{center}
\end{table}

\textbf{Correlation Loss:} To show how the proposed Correlation Loss (CL) contributes to our VQA task, the performances of DCVQE with different weight settings of CL and L1 loss terms are reported in Table \ref{clloss_alpha}. The best performance can be identified when L1 Loss weight $\alpha$ reaches 0.7 and CL weight $\beta$ reaches 0.3. Here we note that our CL does not intend to play a critical role in model optimization since it tries to describe the relative order relationships between videos, particularly for those videos with similar quality scores, so the overall improvement by adding this CL term may not look very significant. Overall, we confirm that our model can achieve better performance with the introduction of CL (Eq. \ref{eq4}). The rest experiments are conducted with the setting of $\alpha=0.7$ and $\beta=0.3$ if not specified.

\setlength{\tabcolsep}{3pt}
\begin{table}\footnotesize
\begin{center}
\caption{Component-by-component comparisons of several key parts in our DCVQE on KoNViD-1k dataset. Here we note that L1 loss is a default setting for all tests. Rows 3 and 4 apply AVG pooling (AVG) and $Transformer_C$ ($Trans_C$) respectively to conduct the ``conquer" operation. ``TR 15"s in rows 5, 6, 7, and 8 mean the temporal ranges of sequence masks are set to 15 for the tests.}
\label{CbyCAblation}
% with different test data splitting operations
\begin{tabular}{clccc}
\hline
% & \multicolumn{3}{c}{TVSum}\\
Row \# & Components & SRCC & PLCC & RMSE \\
\hline 

1 & Vanilla-Transformer   & 0.7988 & 0.7994 & 0.3842  \\
2 & Divide Only	  & 0.7987 & 0.8032 & 0.3723 \\
3 & Divide + AVG	  & 0.8025 & 0.8140 & 0.3644   \\
4 & Divide + $Trans_C$	  & 0.8031 & 0.8121 &  0.3608\\
5 & Divide + AVG + TR 15 & 0.8181 & 0.8267 & 0.3474 \\
6 & Divide + $Trans_C$ + AVG + TR 15     & 0.8278 & 0.8289 & 0.3453 \\
7 & Divide + $Trans_C$ + AVG + TR 15 + PW-RL & 0.8307 & 0.8326 & 0.3542 \\
8 & Divide + $Trans_C$ + AVG + TR 15 + CL & 0.8382 & 0.8375& 0.3515\\
\hline
\end{tabular}
\end{center}

\end{table}
\setlength{\tabcolsep}{1.3pt}

\textbf{Component by Component Ablation:} We further stack the key parts in our DCVQE component-by-component to analyze their impacts individually and gradually. To additionally study the effectiveness of our new CL term, a well-known pairwise ranking loss (PW-RL) \cite{ranknent}, which is similar to our proposal, is also involved in our experiments. As shown in Table \ref{CbyCAblation}, compared with Vanilla-Transformer (row 1), only introduces the Divide operation (row 2)
does not impact the system's performance. In row 3 we employ the AVG pooling operator to integrate clip-level embeddings to generate video-level embeddings, and the results show that this simple “conquer” operation can slightly improve the performance. In row 4 the conquer operation is switched to the proposed $Transformer_C$, where similar correlation scores with a lower RMSE are presented (compared with row 3). The temporal range of sequence mask 15 is introduced to tests in rows 5 and 6, where we can see that this change apparently benefits the system. Furthermore, the 
losses CL and PW-RL are compared directly in rows 7 and 8. The results confirm that the proposal CL provides the best performance for our VQA task.

\subsection{Comparison with the State-of-the-art Methods}
We compare our method with 13 state-of-the-art VQA methods, which can be roughly divided into NR-IQA based methods: BRISQUE \cite{AMittal_BRISQUE}, GM-LOG \cite{WXue_GMLOG}, HIGRADE \cite{HIGRADE}, FRIQUEE \cite{FRIQUEE}, CORNIA \cite{CORNIA}, HOSA \cite{HOSA}, Koncept-512 \cite{KONIQ_10K}, PaQ-2-PiQ \cite{PAQ2PIQ}, and NR-VQA based methods: V-BLIINDS \cite{VIDEO_BLINDS}, TLVQM \cite{TLVQM}, VSFA \cite{VSFA}, VIDEVAL \cite{VIDEVAL} and RAPIQUE \cite{RAPIQUE}. For NR-IQA based methods, we abstract one frame feature per second and regard the average value of the features as the video representation to train a support vector regressor (SVR) for prediction \cite{RAPIQUE,VIDEVAL,PATCHVQ}. 

Test results are reported in Table \ref{comparation} and the best solution for each evaluation metric is marked in bold. 
Since it is hard to reproduce VIDEVAL and RAPIQUE methods, in Table \ref{comparation} we directly cite the results given in \cite{VIDEVAL,RAPIQUE} (marked as “*”). Different from our default 60-20-20 data splitting setting, these two methods used the random 80-20 strategy to split data for training and testing. For a fair comparison, we train an additional version of the DCVQE model under the same 80-20 data splitting setting. Test results of this additional DCVQE model are marked as “\dag”. As seen from this table, our model (DCVQE\dag) significantly outperforms the current top method RAPIQUE on KoNViD-1K dataset by 4.37\% SRCC and 2.45\% PLCC, and greatly exceeds the second place method VIDEVAL on YouTube-UGC dataset by 5.62\% SRCC and 5.67\% PLCC. TLVQM method performs best on LIVE-VQC dataset. Looking into this dataset, we can see that most videos are captured using mobile devices, thus camera motion blurring is one of the common distortions affecting the video quality. TLVQM introduces several hand-crafted motion-related features to handle this issue, so it achieves the best performance, especially compared with all the deep learning based solutions. Nevertheless, besides TLVQM, our proposal performs the best among the remaining. The evaluation on the All-Combined dataset of KoNViD-1k, LIVE-VQC, YouTube-UGC also confirms the robustness of our model. As shown in the column “All Combined”, our method (DCVQE\dag) surpasses the second place method (RAPIQUE) by 2.84\% SRCC and 1.12\% PLCC.

\setlength{\tabcolsep}{1.1pt}
\begin{table} \tiny
\begin{center}
\caption{Performance comparisons with the state-of-the-art methods. 
%\zt{The “*” marked entries denote that results are collected from the original papers, where the experiments are conducted under 80\% training and 20\% testing. The “\dag” marked entries denote that the experiments followed 80\%-20\% settings.}
}
\label{comparation}
\begin{tabular}{lcccccccccccc}
\hline
DATASET & \multicolumn{3}{c}{KoNViD-1k} & \multicolumn{3}{c}{LIVE-VQC} & \multicolumn{3}{c}{YouTube-UGC} & \multicolumn{3}{c}{All Combined} \\
\cline{2-13}
%\cmidrule(l){2-4}  \cmidrule(l){5-7} \cmidrule(l){8-10} \cmidrule(l){11-13}  
MODEL & SRCC & PLCC & RMSE & SRCC & PLCC & RMSE & SRCC & PLCC & RMSE & SRCC & PLCC & RMSE \\
\hline
BRISQUE & 0.6567 & 0.6576 & 0.4813 & 0.5925 & 0.6380 & 13.100 
        & 0.3820 & 0.3952 & 0.5919 & 0.5695 & 0.5861 & 0.5617 \\ 
GM-LOG	& 0.6578 & 0.6636 & 0.4818 & 0.5881 & 0.6212 & 13.223
        & 0.3678 & 0.3920 & 0.5896 & 0.5650 & 0.5942 & 0.5588 \\ 
HIGRADE	& 0.7206 & 0.7269 & 0.4391 & 0.6103 & 0.6332 & 13.027
        & 0.7376 & 0.7216 & 0.4471 & 0.7398 & 0.7368 & 0.4674 \\
FRIQUEE	& 0.7472 & 0.7482 & 0.4252 & 0.6579 & 0.7000 & 12.198
        & 0.7652 & 0.7571 & 0.4169 & 0.7568 & 0.7550 & 0.4549 \\ 
CORNIA  & 0.7169 & 0.7135 & 0.4486 & 0.6719 & 0.7183 & 11.832 
        & 0.5972 & 0.6057 & 0.5136 & 0.6764 & 0.6974 & 0.4946 \\ 
HOSA    & 0.7654 & 0.7664 & 0.4142 & 0.6873 & 0.7414 & 11.353
        & 0.6025 & 0.6047 & 0.5132 & 0.6957 & 0.7082 & 0.4893 \\ 
KonCept-512 & 0.7349&0.7489&0.4260 & 0.6645 & 0.7278 & 11.626
        & 0.5872 & 0.5940 & 0.5135 & 0.6608 & 0.6763 & 0.5091 \\ 
PaQ-2-PiQ & 0.6130&0.6014 & 0.5148 & 0.6436 & 0.6683 & 12.619
        & 0.2658 & 0.2935 & 0.6153 & 0.4727 & 0.4828 & 0.6081 \\ 
\hline     
V-BLINDS   & 0.7101 & 0.7037 & 0.4595 & 0.6939 & 0.7178 & 11.765
          & 0.5590 & 0.5551 & 0.5356 & 0.6545 & 0.6599 & 0.5200  \\ 
TLVQM      & 0.7729 & 0.7668 & 0.4102 & \textbf{0.7988} & \textbf{0.8025} & \textbf{10.145}
          & 0.6693 & 0.6590 & 0.4849 & 0.7271 & 0.7342 & 0.4705  \\ 
%(8-2)VSFA       & 0.7911 & 0.7994 & 0.3880 & 0.7282 & 0.7713 & 10.624
%          & 0.7868 & 0.7920 & 0.3962 & 0.8041 & 0.8151 & 0.4032  \\
VSFA  & 0.7728 & 0.7754 & 0.4205 & 0.6978 & 0.7426 & 11.649
          & 0.7611 & 0.7500 & 0.4269 & 0.7690 & 0.7862 & 0.4253 \\
VIDEVAL*	   & 0.7832* & 0.7803* & 0.4026* & 0.7522* & 0.7514* & 11.100*
          & 0.7787* & 0.7733* & 0.4049* & 0.7960* & 0.7939* & 0.4268* \\ 
RAPIQUE*    & 0.8031* & 0.8175* & \textbf{0.3623}* & 0.7548* & 0.7863* & 10.518*
          & 0.7591* & 0.7684* & 0.4060* & 0.8070* & 0.8279* & 0.3968* \\ 
DCVQE\dag      & \textbf{0.8382}\dag & \textbf{0.8375}\dag & \textbf{0.3515}\dag & 0.7620\dag & 0.7858\dag & 10.549\dag
          & \textbf{0.8225}\dag & \textbf{0.8172}\dag & \textbf{0.3770}\dag & \textbf{0.8299}\dag & \textbf{0.8372}\dag & \textbf{0.3824}\dag \\ 
DCVQE     & \textbf{0.8206} & \textbf{0.8224} & 0.3671 & 0.7479 & 0.7648 & 11.599 & \textbf{0.8069} & \textbf{0.8050} & \textbf{0.3974} & \textbf{0.8239} & \textbf{0.8350} & \textbf{0.3914} \\
\hline  
\end{tabular}
\end{center}
\end{table}
\setlength{\tabcolsep}{1.2pt}

\begin{figure*} 
\begin{center}
\includegraphics[width=0.89\linewidth]{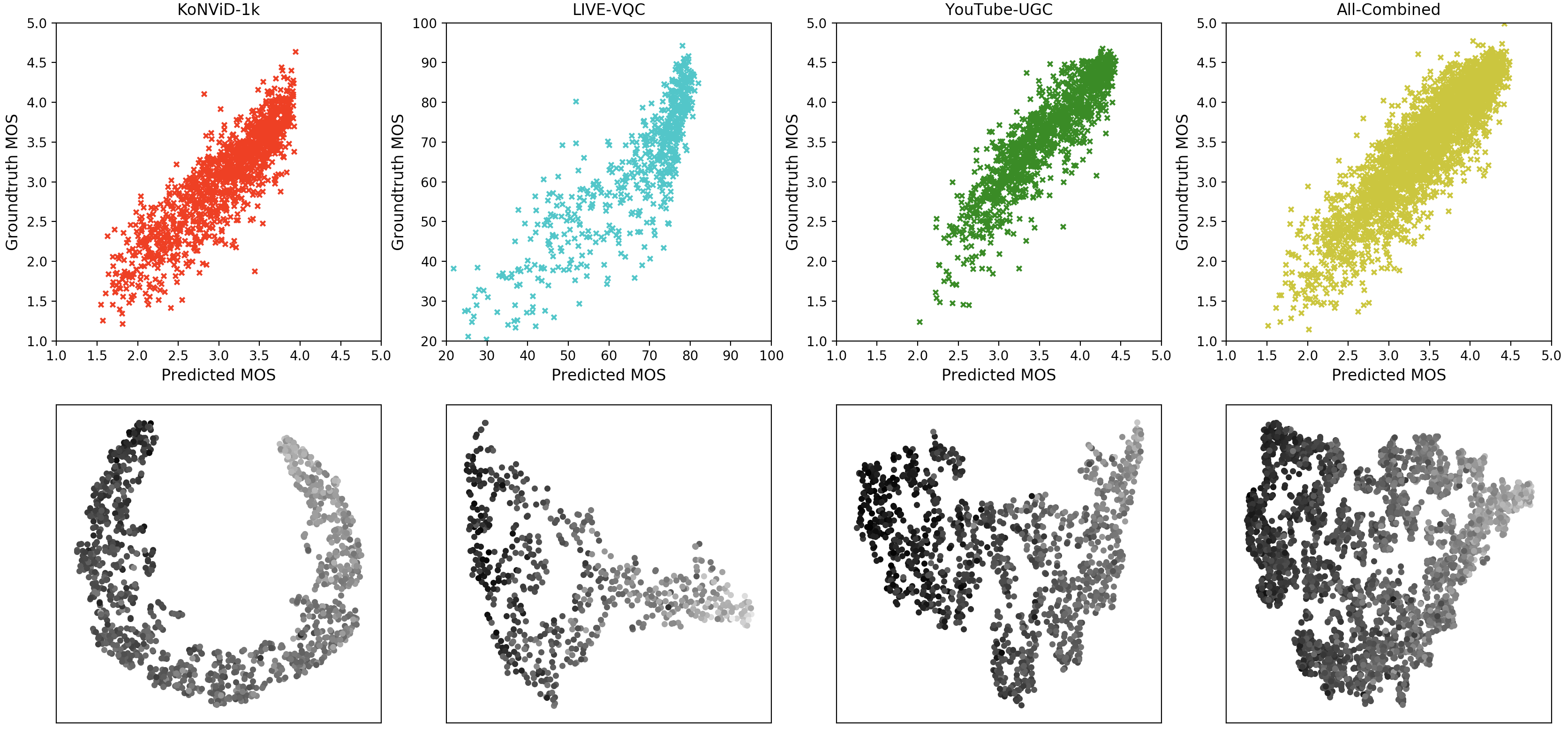}
\end{center}
   \caption{Visualize test results. Top figures: Video quality predictions versus their ground truth MOS scores; Bottom  figures: dimension-reduced video-level QE colored by ground truth MOS scores, where we apply t-SNE algorithm \cite{TSNE} to reduce the dimension of the embeddings to 2 for display.}
\label{fig:tsne}
\end{figure*}

To demonstrate how well our model is learned from a given dataset, in Fig. \ref{fig:tsne} we also visualize the correlation between the video quality predictions and their ground truths (top row), as well as the video-level QE learned by the “conquer” operation of the last DCTr layer (bottom row). From the top row, we can see a strong correlation between our predictions and their ground truths. The figures in the bottom row plot the dimension-reduced video-level QE, where the gray value of each point is assigned with the normalized ground truth MOS score of the corresponding video sample, that is, the darker a point, the greater its corresponding MOS score. From these figures, we can observe that the points with higher MOS scores scatter on one side while the points with lower MOS on the other. It indicates that the separability of quality for those video samples has been significantly enhanced by using our DCVQE model.

\setlength{\tabcolsep}{3pt}
\begin{table}\footnotesize
\begin{center}
\caption{Performance comparisons on the large-scale LSVQ dataset.}
\label{LSVQ}
% with different test data splitting operations
\begin{tabular}{lcccc}
\hline
% & \multicolumn{3}{c}{TVSum}\\
DATASET & \multicolumn{2}{c}{Test} & \multicolumn{2}{c}{Test-1080p} \\
\cline{2-3} \cline{4-5} 
MODEL & SRCC & PLCC & SRCC & PLCC \\
\hline
BRISQUE   & 0.579 & 0.576 & 0.497 & 0.531  \\
TLVQM	  & 0.772 & 0.774 & 0.589 & 0.616  \\
VIDEVAL	  & 0.794 & 0.783 & 0.545 & 0.554  \\
VSFA	  & 0.801 & 0.796 & 0.675 & 0.704  \\
PVQ	      & 0.827 & 0.828 & 0.711 & 0.739  \\
DCVQE     & \textbf{0.836} & \textbf{0.834} & \textbf{0.727} & \textbf{0.758}  \\
\hline
\end{tabular}
\end{center}

\end{table}
\setlength{\tabcolsep}{1.3pt}

Additionally, we compare our model with several others on the recently released large-scale LSVQ dataset. The PVQ model \cite{PATCHVQ}, which was proposed along with this dataset, is also involved in our evaluation. Test results listed in Table \ref{LSVQ} confirm the excellent performance of our model again.

\section{Conclusions}
% \lorem{3}

Inspired by our observation on the actions of human annotation, in this paper, we propose a new Divide and Conquer Transformer (DCTr) architecture to extract the video quality features for NR-VQA. Starting from extracting the frame-level quality embeddings (QE) of an input video, two types of Transformers are introduced to extract the clip-level QE and video-level QE progressively in our DCTr layer. Through stacking several DCTr layers and topping with a regressor, a hierarchical model, named Divide and Conquer Video Quality Estimator (DCVQE) is constructed to predict the quality score of the input video. We also put forward an additional correlation loss regarding the order relationship among the training data to guide the training. Experiments confirm that our proposal outperforms most other methods. What is more, our model is purely deep learning based, compared with other top methods where NSS/NVS features are more or less needed, so we believe our proposal is more practical.

\bibliography{egbib}

\title{DCVQE: A Hierarchical Transformer for Video Quality Assessment (Supplementary Material)}
\author{Zutong Li \and Lei Yang }
\institute{Weibo R\&D Limited, USA\\
\email{\{zutongli0805, trilithy\}@gmail.com}}
% \institute{Weibo R\&D Limited, USA}

\maketitle   
\section{Selection of the CNN Backbone}
We test two series of architectures based on ResNet and ViT with different settings. As Table \ref{table:IQA_Table} shows, we determine that deeper structure results in worse system performance, and partially fine-tuned backbones usually work better than fully fine-tuned ones. For example, the fully fine-tuned ResNet-18$_{full}$ outperforms its corresponding deeper versions ResNet-50$_{full}$ and ResNet-101$_{full}$, while the partially fine-tuned ResNet-50$_{partial}$ shows a significant improvement over its fully fine-tuned one ResNet-50$_{full}$. The same conclusion can be drawn by analyzing the ViT results. Also, we incorporate an attention-based
IQA architecture PHIQNet as the feature extractor to our model. Table \ref{table:MM21} shows the performance comparisons of two models (with different backbones). As seen, the partially fine-tuned ResNet-50 and PHIQNet contribute similarly to our task. For a fair comparison with previous work, as well as the tradeoff between model complexity and performance, we select the partially fine-tuned ResNet-50 as the CNN backbone to construct our DCVQE model.

\begin{table} \footnotesize
\begin{center}
\caption{Performance comparisons of two series of architectures based on ResNet and ViT under full and partial fine-tuning strategies on KoNViD-1K dataset. Here ViT-B16/32 represents ViT base model with 16*16/32*32 input patch size, ViT-L32 represents ViT large model with 32*32 input patch size.}
\label{table:IQA_Table}
\begin{tabular}{lcccc}
\hline
% & \multicolumn{3}{c}{TVSum}\\
  Models & SRCC & PLCC &  KRCC & RMSE\\
\hline
 ResNet-18$_{full}$ &     0.8893 & 0.8798 & 0.6955 & 0.2466   \\
 ResNet-18$_{partial}$ &    0.8888 & 0.8813 & 0.6979 & 0.2494   \\
 ResNet-50$_{full}$ &     0.8507 & 0.8407 & 0.6457 & 0.2889   \\
 ResNet-50$_{partial}$ &    0.9058 & 0.8933 & 0.7168 & 0.2308   \\
 ResNet-101$_{full}$ &    0.8511 & 0.8317 & 0.6365 & 0.2847   \\
 ResNet-101$_{partial}$ &   0.9075 & 0.8962 & 0.7166 & 0.2278   \\ 
 ViT-B16$_{full}$ & 0.8620 & 0.8818 & 0.6759 & 0.3570 \\
 ViT-B16$_{partial}$ & 0.7786 & 0.8103 & 0.5849 & 0.3587 \\
 ViT-B32$_{full}$  &  0.7716 & 0.8066 & 0.5788 & 0.3405 \\
 ViT-B32$_{partial}$ & 0.7639 & 0.8038 & 0.5707 & 0.3878 \\
 ViT-L32$_{full}$ & 0.7881 & 0.8131 & 0.5905 & 0.3267 \\
 ViT-L32$_{partial}$ & 0.8406 & 0.8708 & 0.6516 & 0.2786 \\
\hline
\end{tabular}
\end{center}
\end{table}

\begin{table} \footnotesize
\begin{center}
\caption{Performance comparisons of two models with ResNet-50 and PHIQNet feature extraction backbones. The tests are conducted on KoNViD-1K dataset.}
\label{table:MM21}
\begin{tabular}{cccc}
\hline
% & \multicolumn{3}{c}{TVSum}\\
  Models & SRCC & PLCC  & RMSE\\
\hline

 ResNet-50$_{partial}$ + DCVQE & 0.8382 & 0.8375& 0.3515 \\
 PHIQNet + DCVQE &    0.8376 & 0.8313 & 0.3599  \\
\hline
\end{tabular}
\end{center}
\end{table}

\section{Optimal Number of DCTr Layers}
We conduct the ablation study on KoNViD-1K dataset to find out the optimal number of DCTr layers to construct our DCVQE model. As listed in Table \ref{table:layer}, only one DCTr layer does not adequately solve the VQA problem, while stacking 3 DCTr layers significantly increases the performance. Further increases in layers does not improve the performance. As a result, we set 3 as the optimal number of DCTr layers for our model.

\begin{table} \footnotesize
\begin{center}
\caption{Performance comparisons of the different numbers of DCTr layers.}
\label{table:layer}
\begin{tabular}{cccc}
\hline
 Layer \# & SRCC & PLCC  & RMSE\\
\hline
 1 & 0.7954 & 0.8013& 0.3688 \\
 3 & 0.8382 & 0.8375& 0.3515 \\
 5 & 0.8346 & 0.8305& 0.3592 \\
 7 & 0.8350 & 0.8332& 0.3527 \\
\hline
\end{tabular}
\end{center}
\end{table}

\section{More Studies on the Proposed Correlation Loss}
\begin{figure}[t]
\begin{center}
   \includegraphics[width=.89\columnwidth]{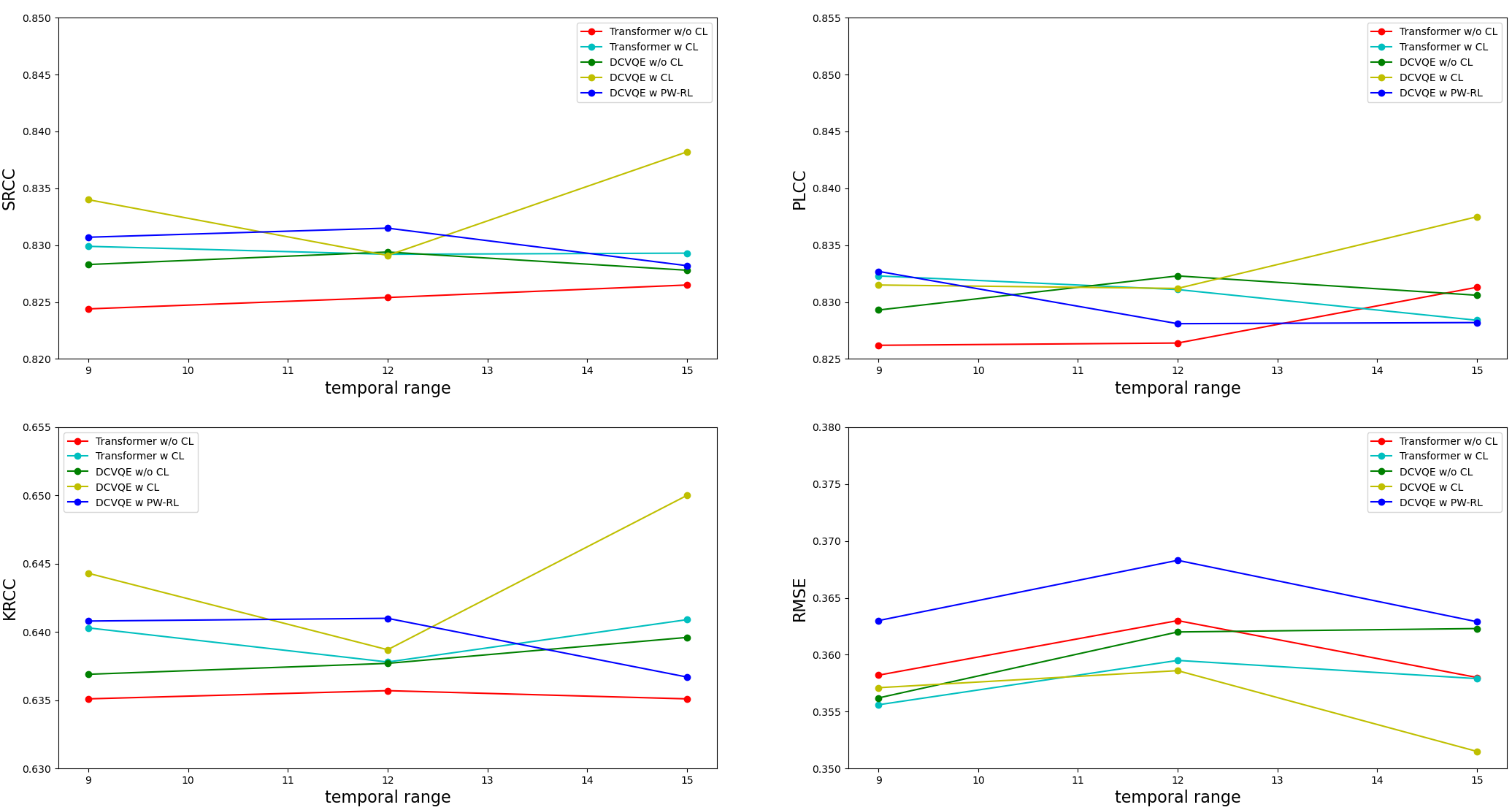}
\end{center}
   \caption{Performance comparisons among ‘with correlation loss’ (w CL), ‘without correlation loss’ (w/o CL), and ‘with pairwise ranking loss’ (w PW-RL) under different models and settings on KoNViD-1K dataset.}
\label{fig:CLLOSS}
\end{figure}

To find out how the proposed correlation loss additionally helps to improve NR-VQA, we apply the proposed losses to train the baseline Transformer and DCVQE models, respectively. The well-known pairwise ranking loss (PW-RL) is also involved in our study. Learned from subsection 4.4 of the paper, the best performance can be achieved with a temporal range selected from 9 to 15, so we only conduct the experiments under 3 different range settings of 9, 12, and 15. The test results are shown in Fig. \ref{fig:CLLOSS}, where we can see that no matter which architecture and temporal range are selected, the introduction of our correlation loss can consistently help to improve VQA performance. The PW-RL is also comparably well to optimize our DCVQE model. However, its solution reaches the highest RMSE. The reason is that the PW-RL will be converted to cross-entropy loss for training so that the optimization strength might be too strong for pairs with wrong ranking orders but small Mean Opinion Score (MOS) differences. Fortunately, our correlation loss can better handle this situation because both ranking orders and MOS differences are considered. 

Additionally, to show how the proposed correlation loss and architecture benefit real VQA tasks, we provide MOS prediction results of 4 KoNViD-1k sample videos in Table \ref{table:example_video}. From this table, we can see that (1) DCVQE$_{cl}$ maintains the order relation among the samples but DCVQE$_{l1}$ and Vanilla-Transformer fail, and (2) both the Mean Absolute Errors (MAEs) of DCVQE$_{cl}$ and DCVQE$_{l1}$ are  lower than that of Vanilla-Transformer thanks to the new hierarchical architecture of DCVQE.

\begin{table} \footnotesize
\begin{center}
\caption{MOS prediction results of 4 KoNViD-1k samples: DCVQE$_{cl}$ is trained with proposed loss (Eq. 5 of the paper); DCVQE$_{l1}$ is trained with L1 loss. }
\label{table:example_video}
\begin{tabular}{ccccc}
\hline
% & \multicolumn{3}{c}{TVSum}\\
  Video Id & Ground Truth & DCVQE$_{cl}$ & DCVQE$_{l1}$ & Vanilla-Transformer\\
\hline
5319047612&	1.35&	1.95&	2.00&	1.99\\
4265470174&	1.56&	1.96&	1.96&	1.97\\
3521396571&	3.54&	3.56&	3.58&	3.72\\
12893008605&	3.55&	3.58&	3.55&	3.68\\

\hline
MAE& - &0.26&	0.27&	0.34\\
\hline
\end{tabular}
\end{center}
\end{table}

\section{Computational Cost Analysis}
Compared with the Vanilla-Transformer, our DCVQE model has a lower computational cost. For example, to calculate the attention weights for one single frame, the time complexity of the Vanilla-Transformer is $O(DN)$, while that of our DCVQE is $O(D*\frac{N}{C})$ because an input video will be split into a number of clips for processing (where $D$ denotes the dimension of feature, $N$ is the total frame size and $C$ is the clip number).

% \section{One More Thing}
% If you are interested in our code, please feel free to contact us. The code can only be used for academic research.

\bibliographystyle{splncs04}
\end{document}